\documentclass[11pt]{article}
\usepackage{coling2020}
\usepackage{times}
\usepackage{url}
\usepackage{latexsym}
\usepackage{color,hyperref}
\usepackage{tabu}

\colingfinalcopy 

\title{Mining Coronavirus (COVID-19) Posts in Social Media}

\author{Negin Karisani \\
  Purdue University \\
  {\tt nkarisan@purdue.edu} \\\And
  Payam Karisani \\
  Emory University \\
  {\tt payam.karisani@emory.edu} \\}

\date{}

\begin{document}
\maketitle
\begin{abstract}
    World Health Organization (WHO) characterized the novel coronavirus (COVID-19) as a global pandemic on March 11th, 2020. Before this and in late January, more specifically on January 27th, while the majority of the infection cases were still reported in China and a few cruise ships, we began crawling social media user postings using the Twitter search API. Our goal was to leverage machine learning and linguistic tools to better understand the impact of the outbreak in China. Unlike our initial expectation to monitor a local outbreak, COVID-19 rapidly spread across the globe. In this short article\footnote{This paper is a short version of a longer study.} we report the preliminary results of our study on automatically detecting the positive reports of COVID-19 from social media user postings using state-of-the-art machine learning models.
\end{abstract}

\section{Introduction and Motivation} \label{sec:intro}

According to a tally by Johns Hopkins University 566,269 persons are tested positive and 25,423 persons have died around the globe as of today, March 27th. Approximately a third of the world's population is impacted by COVID-19. The United States became the epicenter of the virus pandemic on March 26th--yesterday--and New York City with 23,112 confirmed cases is the epicenter of the US outbreak. The US House passed a \$2 trillion stimulus bill to combat the negative impact of COVID-19 on the country's economy. Despite the devastating global impact of COVID-19, WHO has announced that the current pandemic would be the first pandemic in human history that could be controlled.

The impact of COVID-19 on societies is unprecedented. Numerous countries in Asia and the EU, including Iran, Italy, and Spain are under a lock-down. In the US, states such as California and New York are experiencing the same situation. People are ordered to stay home, and are encouraged to practice social distancing. Psychologists advise the residents of the affected areas to practice certain routines to maintain their mental well-being. With people staying at home more often, the role of the internet, as a means of communication, has become even more critical. For instance, NextDoor, a hyperlocal social network, recently announced that the daily rate of its active users has increased by~80\%. 

It has long been known that social networks are effective media for public health monitoring. Despite the well-understood limitations and biases present in the conclusions drawn from social media data \cite{biases}, they are proven to be invaluable resources \cite{survey-1}. In this article, we report the preliminary results of our study on automatically mining the user postings related to COVID-19 on Twitter. Our goal is to find the extent in which machine learning models can distill the user generated data. As pointed out by previous studies \cite{wespad}, this can facilitate the related institutions' responses to the outbreak. In the next section, we focus on automatically detecting the positive reports of COVID-19 infections in the data that we have been collecting since January 27th, 2020.

\section{Dataset}

We started collecting Twitter data on January 27th, 2020. As of March 26th, we have collected 5,621,048 tweets. We used the Twitter search API to crawl the data, and our search keywords were initially ``coronavirus'' and ``corona virus''. On mid-March we added the keywords ``COVID-19'' and ``COVID 19'' to the search criteria. We only collected the English user postings, and omitted retweets and replies.

In order to evaluate the machine learning models we also manually inspected the data, and used stratified sampling to construct a dataset. Thus, for the tweets posted in February we randomly selected between 100 and 300 examples a day and collected the total of 6,090 tweets. This set of tweets constitute our training set. Additionally, we also randomly selected a set of 200 examples a day between March 3rd and March 12th, and collected the total of 2,000 tweets, which constitute our test set. We hired an annotator and annotated the data based on the following criteria: 1) If a tweet mentions individuals infected with COVID-19, and also explicitly or implicitly contains a time reference then it was labeled positive. 2) If a tweet does not mention any individual or lacks a time reference it was labeled negative. To validate the quality of the labels we hired a second annotator and randomly relabeled 10\% of the tweets. The inter-agreement between the two annotators was 0.70 based on Cohen Kappa coefficient, which indicates a substantial correlation \cite{kappa}. Table \ref{tbl-stat} summarizes the training and test sets, and their corresponding percentage of the positive and negative examples. In the next section, we discuss the methods that we implemented and report their results.

\begin{table}
\centering
\begin{tabu}{|p{0.5in} |p{0.5in} |p{0.5in} |[1pt] p{0.5in} |p{0.5in} |p{0.5in} |}
\cline{1-6} \multicolumn{3}{|c|[1pt]}{{\footnotesize \textbf{Training}}}  &
 \multicolumn{3}{c|}{{\footnotesize \textbf{Test}}}  \\ \hline 
{\footnotesize Count} & 
{\footnotesize Negative} &
{\footnotesize Positive} & 
{\footnotesize Count} & 
{\footnotesize Negative} & 
{\footnotesize Positive} \\ \hline
6090 & 90.8\% & 9.2\% & 2000 & 94.1\% & 5.9\% \\ \hline 
\end{tabu}
\caption{The number, and the percentage of the positive and negative tweets in the training and test sets.} \label{tbl-stat}
\end{table}

\section{Experiments}

We begin this section by describing the methods that we implemented, then we briefly discuss the training procedure, and finally report the results.

\subsection{Methods Compared}

We included seven methods in our experiments. One classic generative model (Naive Bayes), one classic discriminative model (Logistic Regression), one widely used neural network model (fasttext), and four models based on the state-of-the-art model Bidirectional Encoder Representations from Transformers~(BERT). Below we briefly describe each one.

\noindent\textbf{\textit{NB}.} We included the Naive Bayes classifier. We incorporated the MALLET implementation \cite{mallet} of this classifier, and used the tweet unigrams and bigrams as features.

\noindent\textbf{\textit{LR}.} We included the Logistic Regression as the discriminative counterpart of Naive Bayes. We incorporated the MALLET implementation, and again used the tweet unigrams and bigrams as features.

\noindent\textbf{\textit{Fasttext}.} We included the neural model introduced in \cite{fasttext}. This model is a shallow wide network, capable of updating the input word embeddings during the training. We used the pretrained word2vec vectors \cite{word2vec} as input features. The learning rate was empirically set to 0.5, and the window size was set to 2.

\noindent\textbf{\textit{BERT-BASE}.} We included the state-of-the-art model introduced in \cite{bert}. This model is based on a multi-layer transformer encoder \cite{transformer}. We used the pre-trained base variant, followed by one layer fully connected network. We applied the default model settings recommended in \cite{bert}. We used the pytorch implementation of BERT introduced in \cite{bert-impl}.

\noindent\textbf{\textit{BERT-Twitter}.} Since our classification problem is defined on social media posts, we can expect that a model specifically exposed to the social media language model (through the Masked Language model task) would perform better than regularly pre-trained ones. Thus, we used a corpus of 35 million tweets collected between 2018 and 2019 through the Twitter streaming API to further pre-train \textit{BERT-BASE}. We set the maximum window size to 160 and batch-size to 32--the rest of the settings were set to the default values, as suggested in \cite{bert}--and pre-trained this model for 4.5 million steps--about 5 epochs. 

\noindent\textbf{\textit{BERT-Corona}.} We hypothesized that if a model is already familiar with the contexts in which the word coronavirus is used in, it would perform better during the classification phase. Thus, we used the tweets that we collected between January 27th and March 3rd to pre-train \textit{BERT-BASE}. We pre-trained this model for 400K steps--approximately 5 epochs. The pre-training settings were identical to \textit{BERT-Twitter}.

\noindent\textbf{\textit{BERT-Corona-BiLSTM}.} Even though BERT already utilizes a sophisticated attention mechanism, we still experimented with sequence encoding models. Thus, we used a Bidirectional Long Short Term Memory Network \cite{lstm} on top of \textit{BERT-Corona}, followed by one layer fully connected network. We empirically observed that if we set the size of the hidden dimensions of the BiLSTM to a half of the size of the hidden dimensions of BERT (i.e., 768) we would get the best performance.

\subsection{Experimental Setup}

We trained \textit{Fasttext} for 100 iterations. The models based on BERT, i.e., \textit{BERT-BASE}, \textit{BERT-Twitter}, and \textit{BERT-Corona}, were trained for 2 iterations--these models are already based on a pre-trained model. We trained \textit{BERT-Corona-BiLSTM} for 3 iterations, since it has more parameters than the other BERT based baselines. For training the models, we used the default model optimizers proposed by the references. In none of the experiments we did any text pre-processing. Since there is a randomness in model initialization and drop-out regularization, we carried out all of the neural network experiments for five times. The results reported in the next section are the average over these experiments. 

The task that we defined is a binary classification problem--detecting the positive reports of COVID-19 from Twitter data. Since the class distribution is highly skewed, following the previous studies \cite{metrics}, we report the F1, Precision, and Recall of the models in the positive class. 

\subsection{Results}

Table \ref{tbl-result} summarizes the performance results. We see that the baselines based on the pre-trained models show the best results. The experiments validate our hypothesis about the effectiveness of pre-training BERT on domain specific data. We see that \textit{BERT-Corona} has achieved the best F1 value. By comparing \textit{BERT-Twitter} and \textit{BERT-Corona-BiLSTM} we can also hypothesize that model initialization--through pre-training--can be potentially more effective than increasing model complexity. Even though validating this hypothesis require more comprehensive experiments.

\begin{table}
\centering
\begin{tabular}{|p{1.5in} |p{0.5in} |p{0.5in} |p{0.5in} |}\hline
\textbf{Model} & 
{\footnotesize \textbf{F1}} & 
{\footnotesize \textbf{Precision}} & 
{\footnotesize \textbf{Recall}} \\ \hline
\textit{NB} & 0.109 & \textbf{0.700} & 0.059 \\ \hline 
\textit{LR} & 0.470 & 0.662 & 0.364 \\ \hline 
\textit{Fasttext} & 0.530 & 0.650 & 0.447 \\ \hline 
\textit{BERT-BASE} & 0.645 & 0.589 & 0.715 \\ \hline 
\textit{BERT-Twitter} & 0.666 & 0.613 & \textbf{0.730} \\ \hline 
\textit{BERT-Corona} & \textbf{0.676} & 0.632 & 0.727 \\ \hline 
\textit{BERT-Corona-BiLSTM} & 0.659 & 0.666 & 0.654 \\ \hline 
\end{tabular}
\caption{Average F1, precision, and recall in detecting positive reports of COVID-19 from Twitter data.} \label{tbl-result}
\vspace{-0.5cm}
\end{table}

We believe a robust social media surveillance system can be immensely helpful. Although the results are encouraging, there are still a lot of challenges to be addressed to build such a system for COVID-19. Automatically detecting positive reports, or even following up on the mental well-being of patients through social media posts can greatly enhance the concerned institutions' endeavor to monitor the public health and respond in timely manner.

\section{Conclusions}

In this short article we reported the preliminary results of our study on the capability of machine learning models to distill social media posts related to COVID-19, namely we focused on automatically detecting the positive reports of this illness. We constructed a manually annotated dataset, and showed that state-of-the-art classifiers have encouraging results. Our pre-trained model and unlabeled data can be accessed through our Github \footnote{Available at~\url{https://github.com/nkarisan/Covid19_Research}}. We will also release our labeled data, along a more comprehensive analysis soon.

\bibliographystyle{coling}
\bibliography{coling2020}

\end{document}